 \documentclass[final,3p,times,twocolumn]{elsarticle}

\usepackage{booktabs}
\usepackage{placeins}
\usepackage{graphicx}
\usepackage{amsmath}
\usepackage{color}

\usepackage{amssymb}

 \usepackage{lineno}

\journal{Building and Environment (accepted for publication)}

\begin{document}

\begin{frontmatter}





\title{Window Opening Model using Deep Learning Methods}


\author[label1]{Romana Markovic\corref{cor1}}\ead{markovic@e3d.rwth-aachen.de} \author[label1]{Eva Grintal} \author[label1]{Daniel W\"{o}lki}  \author[label1]{J\'{e}r\^{o}me Frisch} \author[label1]{Christoph van Treeck}

\address[label1]{E3D - Institute of Energy Efficiency and Sustainable Building, RWTH Aachen University, Mathieustr. 30, 52074 Aachen, Germany}
\cortext[cor1]{Corresponding author. Tel.: +49-241-80-25541 ; fax: +49-241-80-22030.}
\begin{abstract}
Occupant behavior (OB) and in particular window openings need to be considered in building performance simulation (BPS), in order to realistically model the indoor climate and energy consumption for heating ventilation and air conditioning (HVAC). However, the proposed OB window opening models are often biased towards the over-represented class where windows remained closed. In addition, they require tuning for each occupant which can not be efficiently scaled to the increased number of occupants. This paper presents a window opening model for commercial buildings using deep learning methods. The model is trained using data from occupants from an office building in Germany. In total, the model is evaluated using almost 20 mio. data points from 3 independent buildings, located in Aachen, Frankfurt and Philadelphia. Eventually, the results of 3100 core hours of model development are summarized, which makes this study the largest of its kind in window states modeling. Additionally, the practical potential of the proposed model was tested by incorporating it in the Modelica-based thermal building simulation. The resulting evaluation accuracy and \textit{F1} scores on the office buildings ranged between 86-89 \% and 0.53-0.65 respectively. The performance dropped around 15 \% points in case of sparse input data, while the F1 score remained high.
\end{abstract}

\begin{keyword}
deep learning \sep neural networks \sep occupant behavior \sep window opening \sep natural ventilation

\end{keyword}

\end{frontmatter}


\section{Introduction}

Window openings were identified to have a high impact on the energy consumed to sustain the desired indoor environmental quality level \cite{fabi2012}. In addition, it is common knowledge that the window states are one of the required information for modeling the natural ventilation in commercial and residential buildings and they are an important part of thermal building simulation \cite{wolf2017}.
\\
However, window openings and closings are a product of the complex combination of physical, comfort and behavioral models of building occupants \cite{hong2013}. As such, the position of operable windows can not be modelled using a physical analytical approach similarly to other physical heat transfer systems in buildings. Therefore, window states are modelled using either stochastic or machine learning approaches. \\
Data driven approaches, including stochastic and machine learning modeling of window states have shown satisfying performance regarding the prediction of the window states. However, they show poor generalization capabilities and low performance when applied to an unknown building or even to a previously unseen user in the same building. As a result, a model fine-tuning for each occupant is required, which results in high computational costs. 
\\
This paper proposes a generic model that identifies window states using a deep feed-forward neural network. Optimal model formulation is conducted using an extensive hyperparameter search and the model is trained using the data from a subset of three monitored offices. The evaluation is conducted using the data from another 49 offices, resulting in approximately 19 mio. evaluation samples. The research questions addressed by this study are the following:
\begin{itemize}
\item{what are suitable multi-layer perceptron architecture and hyperparameters for modeling the window states in commercial buildings?}
\item{could the window opening habits of a large number of occupants be learned using the data from a relatively small (3 out of 52 offices) subset? }
\end{itemize}
 To present the practical potential and limitations of the proposed modeling approach, additional case studies were conducted:
\begin{itemize}
\item{the model is evaluated using the data from an U.S. office building, which was not included in the model training}
\item{the model is evaluated using a sparse data set from an additional office building, where monitoring data for 30 \% of the features were missing}
\item{the developed model is incorporated in a Modelica-based thermal building simulation}
\end{itemize}
The structure of the rest of the paper is as follows: section 2 reflects on the related studies on OB and deep learning for OB modeling; section 3 includes the analysis of the monitoring data  and the identification of the similarities between a number of occupants; the model development and design of additional case-studies is described in sections 4 and 5. Eventually, the evaluation results are presented,  elaborated and summarized in sections 6-8.
\section{Related Research}
\subsection{OB}
The impact of OB on energy consumption has been subject of research in fields of architecture, engineering and social science. In particular, there exists a number of studies that pointed out that the human factor has a significant impact on the consumed energy in buildings \cite{bonte2014}, \cite{mahdavi2008}, \cite{sun2017}, \cite{dewilde2014}. Hong et al. \cite{hong2013} conducted a study with the aim to quantify the impact of OB on energy consumption in private offices. Their findings pointed out, that different behavioral models resulted in between 50 \% under- and 90 \% overestimation of energy consumption, compared to the reference value. Similarly, OB is responsible for a variance in the energy performance of residential buildings. Based on the study of 209 identical households in Denmark, Andersen et al. \cite{andersen2009} showed that there were large differences in the behavior of occupants between individual dwellings. In addition, Cal\'{i} et al. \cite{cali2016} confirmed that OB affects building's energy performance as well as that the occupants' diversity results in a variance of actual consumed energy in identical dwellings. As a conclusion from the presented studies, the human factor is identified as a cause of the gap between the predicted and measured energy consumption in buildings. Resultantly, there was a number of research studies that proposed predictive models of OB in terms of the occupant's manual control of sun shades (\cite{obrien2013}, \cite{motamed2017}), use of air conditioning (\cite{dong2014}, \cite{dong2009}, \cite{peng2018}, \cite{zhao2016}, \cite{dong2014}), plugin loads (\cite{hong2017}, \cite{virote2012}, \cite{zhao2014}) and window openings.

\subsection{Window opening behavior}
The number of studies on predictive models of window openings spiked out compared to further occupant's actions. This may be caused by the high percentage of manually operable windows as a current state of building production and retrofitting, and due to the advances in recent research. Rijal et al. \cite{rijal2007}, \cite{rijal2008} developed adaptive models for predicting the window openings based on indoor and outdoor air temperatures. The proposed modeling approaches were evaluated in numerous later studies (\cite{schweiker2012}, \cite{laurent2017}, \cite{mahdavi2016}, \cite{langevin2015}), and it was shown that the validation results on additional evaluation data sets of comparable order, when compared to the initially presented predictive performance. Haldi and Robinson \cite{haldi2009} made a significant contribution to the field by developing a model of window states based on eight years of monitoring data. They proposed a classification of window status using logistic regression and a hidden Markov model for incorporating the time-series features of window openings. Later double-blind studies (\cite{wolf2017}, \cite{laurent2017}, \cite{haldi2013}, \cite{schweiker2009}), where the original model was validated using an independent data set, pointed out a low model performance and a need for the model retraining. Although the key challenges were already identified in previous studies (\cite{mahdavi2016}, \cite{haldi2013}), they are still an issue in the proposed modeling approaches, namely:
\begin{itemize}
\item{achieving satisfying accuracy for modeling the imbalanced data, since the proportion of closed windows is larger than opened,}
\item{need for building-wise or even occupant-wise parameter search.}
\end{itemize}
In recent publications, there were multiple studies that proposed occupant wise-tuning for the window opening models (\cite{langevin2015}, \cite{markovic2017}). Langevin et al. \cite{langevin2015} proposed an agent-based approach for modeling the occupant behavior in commercial buildings in terms of use of fans, window openings and heaters. Markovic et al. \cite{markovic2017} compared the performance of classification algorithms for window opening modeling. In both studies, there was a larger proportion of correctly identified window states, compared to the building-wise model tuning. Additionally, the proportion of correctly identified window states outperformed the alternative approaches. However, the occupant-wise tuning may not be scalable for the large number of occupants. 
In addition to the above mentioned key challenges, the recent research in the field pointed out that it is essential to represent the human diversity in OB, in order to produce a valid and reliable modeling approach \cite{hong2016}. 
\subsection{OB diversity}
Even though there are advances in the theoretical definition of OB diversity and the attributes that lead to the diversification of human energy-related behavior \cite{doca2014}, \cite{parsons2002}, \cite{andersen2009}, there is little work done on a practical modeling and model applicability.
Feng et al. \cite{feng2016} conducted a large-scale questionnaire of occupant's air conditioning use patterns. They concluded that an unsolved problem regarding modeling the occupants' diversity is labeling the behavioral patterns from a large sample of data and that there is a need for research on practical algorithmic representation of the users' behavioral patterns.
However, the lack of a generic predictive model of OB may be connected to the current state of modeling techniques. Namely, the research on predictive models of OB is strongly grounded on stochastic approaches. Hence, stochastic modeling results in a suppressed occupant diversity \cite{obrien2017}, and thus, there is a need for more generic modeling methods that are suitable for a large number of occupants that are capable to recognize and learn the individual behavioral patterns. Here, deep learning methods are identified as a suitable approach.
\subsection{Deep learning for OB modeling}
The potential of deep learning methods for energy consumption related OB modeling has already been researched by a number of studies \cite{zhao2016}, \cite{coelho2017}, \cite{fan2017}, \cite{kontokosta2017} \cite{kazmi2016}, \cite{kong2017}. Coelho et al. \cite{coelho2017} designed a graphics processing unit (GPU)-based parallel strategy for time-series learning of energy consumption. They concluded that the proposed GPU strategy could be scalable to a large number of time-series for model training, resulting in 45 times faster computations, compared to a single central processing unit (CPU) approach. Zhao et al. \cite{zhao2016} developed a deep recurrent network for detecting the number of occupants in a room. The method uses indoor climate and the information about present heat sources as model input. The results showed that the prediction error is 0.75 \% in case of a three-layer recurrent neural network.
\cite{fan2017} presented a deep learning model for predicting the cooling load for the following 24 hours. They concluded, that the best performance is achieved if the input features were created through feature representation using unsupervised methods. In addition, they pointed out that two hidden layers were sufficient to achieve an optimal performance. Kontokosta and Tull \cite{kontokosta2017} used deep learning methods to predict the energy intensity of 1.1 mio. buildings in New York. The results showed that the electrical energy consumption could be reliably predicted using the data from a relatively small subset of buildings.

\section{Data set}
The introduced model was developed using the data collected at the E.ON Energy Research Center (in further text referred as E.ON ERC) in Aachen, Germany. Building data are collected as a part of a long-term monitoring study, conducted by E.ON ERC on RWTH Aachen University's Building ID 4120. The building itself is an university office building. The available data were logged in a minute-wise frequency between January 1st, 2014 and October 1st, 2015. The data set includes detailed indoor climate, air quality and OB information in 52 single or double occupied offices. For additional information regarding the data preprocessing and the composition of the data set, the reader is referred to the supplementary material, Appendix A.1.

\begin{figure*}[th!]
\centering
 \includegraphics[trim=4cm 0cm 4cm 5cm, width=1\textwidth]{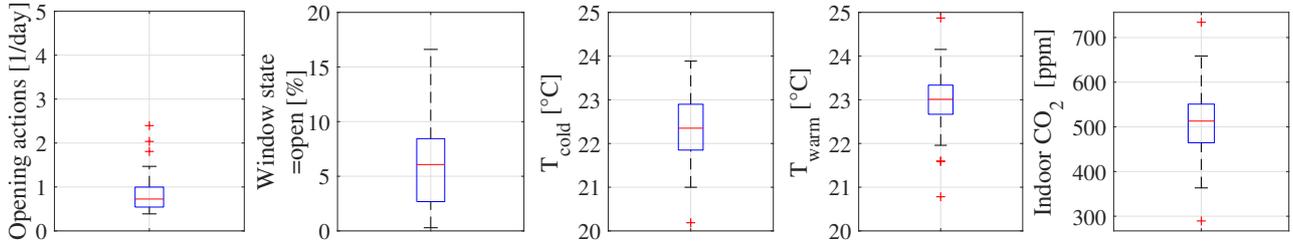}
\caption{Boxplots representing the occupants' actions in terms of window openings and monitored indoor climate in 52 analyzed offices from E.ON ERC Building.}
\label{fig:boxstat_}
\end{figure*}

The habits of the monitored occupants in terms of window opening actions per day, percentage of time where windows were opened, and the measured indoor air temperature on warm and cold days were analyzed. The resulting office-wise distributions are presented in Figure \ref{fig:boxstat_}. Additionally, distribution of the mean indoor $CO_2$ concentration for each office is considered (Figure \ref{fig:boxstat_}), since it reflects the occupant's activity level and occupancy rate. 
\subsection{Occupant segmentation}

Occupants were grouped based on their physiological and behavioral features. For that purpose, the mean indoor air temperature on cold days, mean indoor air temperature on warm days, the $CO_2$ level and the percentage of time at which windows remained opened were analyzed for each monitored office. Here, cold periods were defined as outdoor temperatures below 12 $^{\circ}$C. The behavioral features used for building the groups are analyzed through the variables percentage of time where window sate was logged as “opened”, indoor air temperature on warm days and indoor air temperature on cold days.

\FloatBarrier

\begin{table*}[ht!]

\centering
   \caption{Overview of the used features for the model development and evaluation; comparison of the set bounds for feature scaling and measured values from the E.ON ERC. }
     \begin{tabular}{llrlrlr}
     \toprule
     \multicolumn{1}{l}{Data set} &  & \multicolumn{2}{c} {Defined range} & \multicolumn{2}{c} { E.ON ERC data }    \\
\toprule
      \multicolumn{1}{l}{Indoor climate} & unit& min   &max   & min   &max     \\
     \midrule
     \multicolumn{1}{l}{Hour} & -  & 0  &23    &0  & 23   \\
     \multicolumn{1}{l}{Day of the week} &  -  & 0  & 6    & 0  & 6  \\
     \multicolumn{1}{l}{Presence} & -   &  0 & 1   &0   & 1  \\
          \multicolumn{1}{l}{ $CO_{2}$  } & ppm  & 0  & 2500 & 192  & 1970  \\
     \multicolumn{1}{l}{Relative humidity} & \%  & 0  &   100 & 0.6  & 100  \\
     \multicolumn{1}{l}{Set temperature T1} & $^{\circ}$C  &   18& 26   & 15   & 24  \\
          \multicolumn{1}{l}{Set temperature T2} & $^{\circ}$C  &  18 &26    &   18 & 24     \\
     \multicolumn{1}{l}{Indoor air temperature} & $^{\circ}$C  & -10  & 40   & 9.80    & 31.70  \\
          \multicolumn{1}{l}{Outdoor temperature} & $^{\circ}$C  & -10  &   50 &  2.20  & 45.70   \\
               \toprule
     \multicolumn{1}{l}{Weather data} & unit  & min   &max  & min   &max    \\
\toprule          
     \multicolumn{1}{l}{Timestamp} & Unix time  &1.10e+09  & 1.58e+09   &1.39e+09  & 1.44e+09 \\
     \multicolumn{1}{l}{Avg. temperature} & $^{\circ}$C  & -10  & 40  & -5.41  & 37.49   \\
          \multicolumn{1}{l}{Avg. rel. humidity} & \%   & 0  & 100    & 17.84  &  100  \\
     \multicolumn{1}{l}{Avg. temp. h=-100 cm} & $^{\circ}$C  &  -10 & 40  &4.35    & 18.33    \\
     \multicolumn{1}{l}{Rain droplets total} & - &  0 &  15  & 0 & 10.72  \\
     \multicolumn{1}{l}{Rain droplets volume} &  - & 0  & 0.5   & 0  & 0.20    \\
     
          \multicolumn{1}{l}{Max. wind speed} & m/s  &   0& 28.61   & 0  &24.74    \\
     \multicolumn{1}{l}{Wind direction} & deg   &0   &  360  &  0 &  356.50\\
          \multicolumn{1}{l}{Wind speed} & m/s   & 0  &  28.61   &  0 & 13.20   \\
     \multicolumn{1}{l}{Avg. pressure} & mbar  & 900  &  1100  & 959.39 & 1010.14  \\
     \multicolumn{1}{l}{Global radiation} & W/$m^{2}$   & 0  & 1362   & 0  & 1159  \\
     \multicolumn{1}{l}{Diffuse radiation} & W/$m^{2}$   &0   &  800   &  0  & 485.10 \\
                    \toprule
   \multicolumn{1}{l}{Label} & unit  & min   &max    & min   &max    \\
\toprule  
     \multicolumn{1}{l}{Window position} &-    & 0  & 1  & 0  & 1  \\
     \bottomrule
     \end{tabular}
   \label{tab:data_ebc}
\end{table*}  
\FloatBarrier

Based on the percentage of time where windows were opened, the habits of the monitored occupant in terms of overall duration of time where they kept windows opened in the office was analyzed. Since the occupants had control over the indoor air temperature through the thermostats available in each office, turning on/off mechanical ventilation and through the manually operable windows, it was considered that the resulting indoor air temperature was directly influenced by occupant's actions and their proffered indoor climate. Additionally, the occupants were the main source of the indoor $CO_2$ in the monitored offices. Due to that, measured $CO_2$ values could reflect the occupants' activity and resulting metabolic rate. It was aimed to define the training set as the data subset that belonged to occupants that showed “typical” behavior for the monitored building. The idea behind was that the actions of the "typical" occupants were caused by the same drivers, as the significant proportion of actions performed by the total occupants population. As a result, the common predictors of the window opening actions should be learned by the neural network. For that purpose, the training set consisted of occupants that belonged to each of the three largest clusters.
\begin{figure*}[th!]
\centering
 \includegraphics[trim=0cm  0cm 0cm 3cm, width=\textwidth]{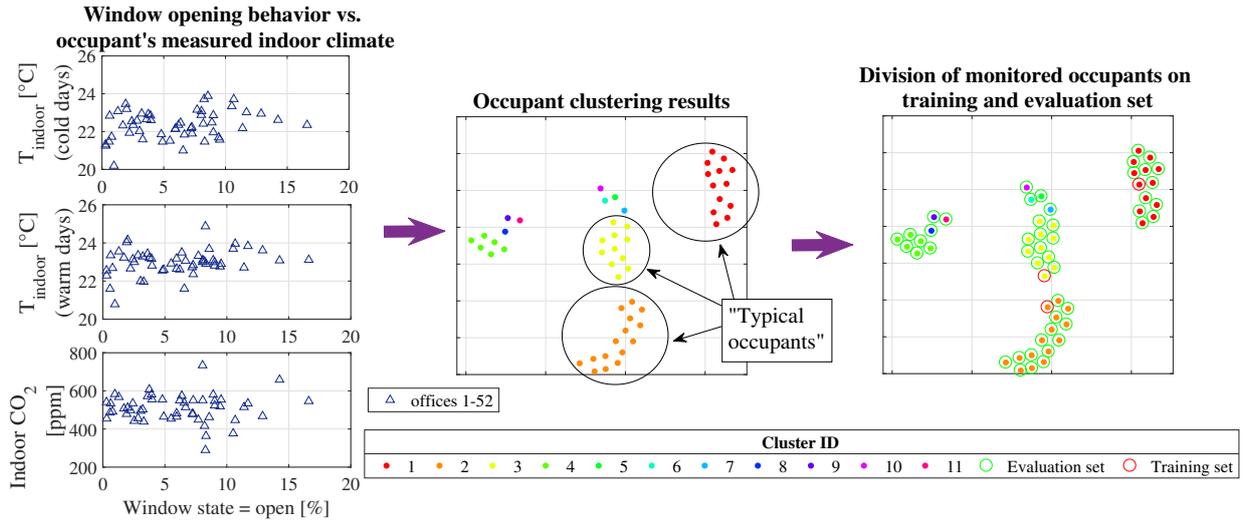}
\caption{Occupant segmentation results. Starting point was the analysis of the dependence between the measured indoor climate and window opening behavior (left hand side). The occupants were clustered based on these dependencies and the 4-dimensional space was projected in 2D using t-SNE visualization algorithm (middle). Finally, the occupants were split into training and evaluation set (right hand side).  }
\label{fig:tsne}
\end{figure*}
\FloatBarrier

These occupants were chosen randomly and their projection in two-dimensional space was not used as one of the criteria for that intra-cluster distinction of occupants from the training and evaluation set. The reason for that was, that the used embedding was a probabilistic approach and the cluster visualization would result in different shaped projection after each repeated visualizations. However, clusters could be spacially separated for each repeated visualization. To present the effect of probabilistic features of t-SNE visualization, several separately produced visualization using the same parameters are provided in the supplementary material, appendix B.
\\
The occupants were grouped based on the above listed features using hierarchical clustering with a bottom up approach, since the number of clusters and the cluster shape were not known a priori. Since the clustering was conducted on 52 data points, where one data point represents an office, the model complexity remained low.  It was iterated over the number of splits and number of clusters until the occupants could be assigned to distinct clusters. The applied clustering was a deterministic approach, so that the repeated clustering of the same data would always result in the same solution. The results of the four dimensional occupants' clustering were projected in 2 dimensions using the t-SNE algorithm \cite{vanmaaten2008}.
As presented in Figure \ref{fig:tsne}, around 75\% of occupants could be grouped into 3 clusters. In the further model development, these clusters will be assumed to represent the average occupants from the monitored commercial buildings.

\subsection{Do different occupants behave similarly?}
\begin{figure}[h!]
\centering
 \includegraphics[trim=2cm  4cm 14cm 1cm, width=0.5\textwidth]{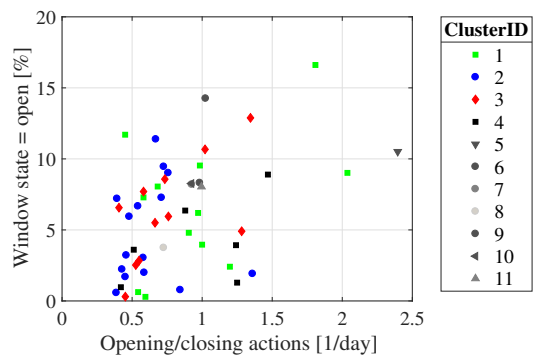}
\caption{Relationship between the proportion of time where windows were opened and mean number of opening actions per day for 52 offices from 11 clusters. }
\label{fig:scatter_actions}
\end{figure}

As presented in previous section, the occupants from the analyzed building were grouped in distinct clusters based on the information stored in physical measurements of indoor climate. In the following step, the frequency of their actions was analyzed together with the proportion of time where windows were opened. As presented in Figure \ref{fig:scatter_actions}, there exist no clear pattern in the occupants frequency of window opening actions that could be associated with the clusters of occupants’ control of indoor environment. The results confirmed the already established conclusions from the related research \cite{andersen2013}, since the actions were driven by different variables, although the occupants from this data set showed similarities in terms of small number of openings per day and the low proportion of time where windows were opened. 

\FloatBarrier

\FloatBarrier
\begin{figure*}[t!]
\centering
 \includegraphics[trim=0cm  0cm 0cm 18cm, width=0.85\textwidth]{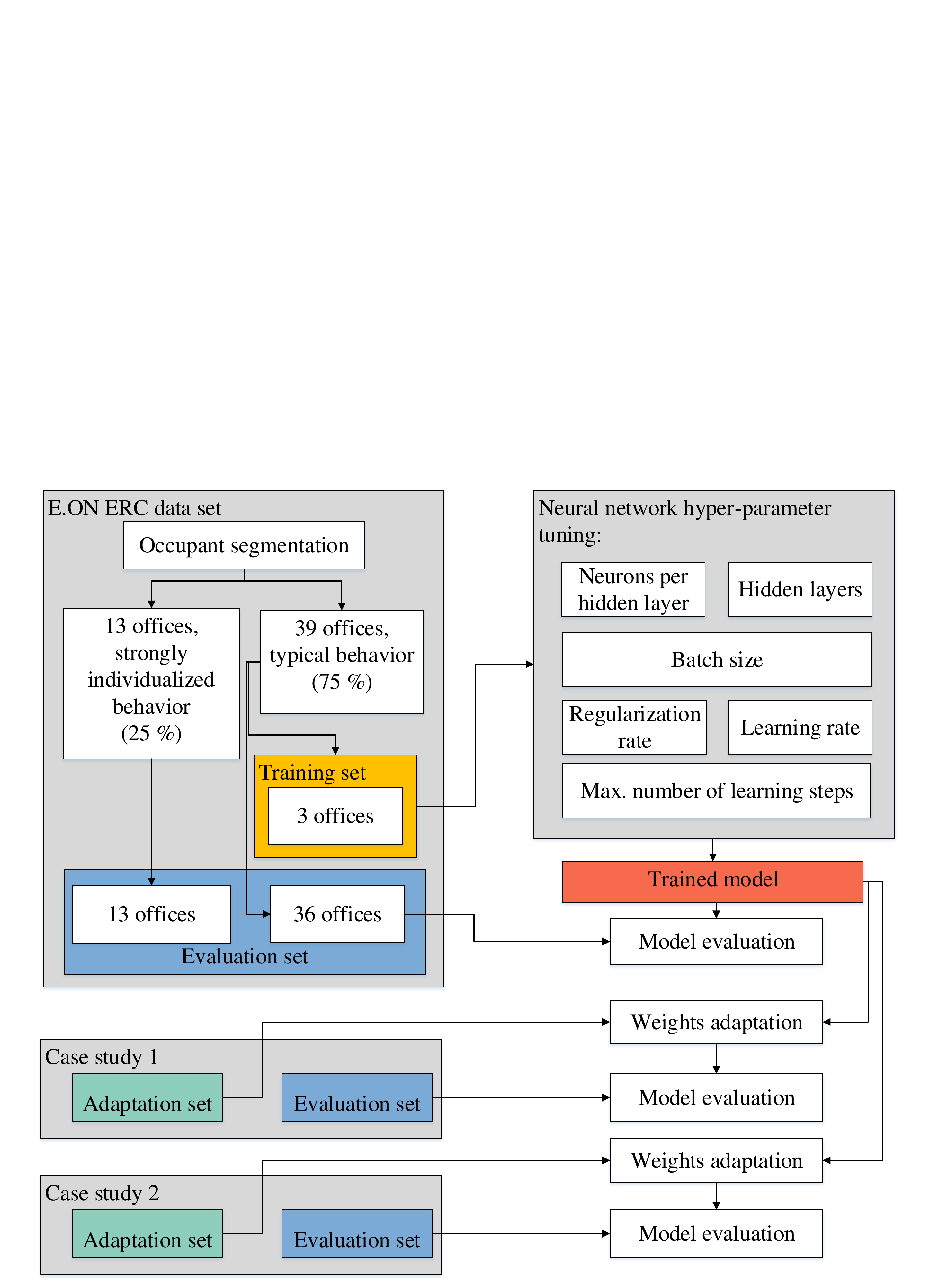}
\caption{Overview of the proposed method for choosing the training data set, model training and evaluation.}
\label{fig:model_overview}
\end{figure*}
\FloatBarrier

\section{Method}
The proposed approach includes the identification of universal offices in terms of adjusted indoor air temperature, activity based on $CO_2$ measurement and the ratio of opened/closed windows. A deep learning based window opening model was trained using the data collected on three universal occupants/offices. The trained model was eventually applied on two additional data sets. Here, the pre-trained weights were adapted by running several further tuning iterations, while no hyperparameter tuning or further calibration was required. An overview of the modeling approach is presented in Figure \ref{fig:model_overview}.

\FloatBarrier
\subsection{Neural network training}
A fully connected feed-forward neural network was trained to identify the window states, using the data collected on the three offices considering the segmentation results presented in the previous section. This resulted in a training set size of approximately 800k data points. The model is evaluated using data from the further 49 offices. During the model training, the investigated parameters include the neural network architecture in terms of number of hidden layers, amount of neurons for each layer, choice of batch strategy, learning rate, impact of regularization parameter and number of iterations.

For the classification of window states in office buildings, it was opted for a fully connected feed-forward neural network, for which an optimal number of hidden layers and neurons for each hidden layer were investigated. A suitable choice of the number of hidden units was necessary in order to find an optimal trade-off between its representation capacity and a use of too large memory in case of complex networks \cite{goodfellow2017}. 
The size of the input layer corresponds to the number of features, while the size of the last layer corresponds to the size of the output. In case of this study, the network consists of 25 neurons in the input layer, corresponding to 22 features from the current time step as well as indoor air temperature, indoor air humidity and $CO_2$ concentration 10 minutes before the current time step. A single neuron in the output layer corresponds to the predicted window state. 

\subsection{Evaluation metrics}
The developed model was evaluated using the previously unseen data from the E.ON ERC data set as well as using a data set collected through two independent monitoring studies.
Firstly, model was evaluated using the data points that were not applied in the training procedure. As a result, the model was evaluated using 19 mio. data points collected on 49 offices located in the same office building in Aachen.
\\
Since the randomized initial weights were defined for the model training, the training and evaluation procedure for the optimal hyperparameter combinations were repeated multiple times, and the final evaluation performance was computed as the mean accuracies of performed iterations.
\\
The model performance was evaluated using the indicators as proposed by \cite{mahdavi2016}, namely confusion matrix (consisting of true positive rate (TRP), true negative rate (TNR), false positive rate (FPR), false negative rate (FNR)), accuracy (ACC), overall fraction of time where windows were opened, mean number of actions per day, open- and closed state median, 25$^{th}$ and 75$^{th}$ percentile and interquartile (IQR, defined as the difference between the latter two values). Here, the number of actions per day was defined as proposed by Mahdavi et al. \cite{mahdavi2016}, namely, number of actions, divided by the cumulated duration of monitoring data in days. Respectively, the proportion of time where the window state was “open” was defined as the number of data points, where window was opened divided by the total number of data points. The opening and closing duration was defined as the time-sequences, where the window state was not changed. Here, only the sequences, where no data point was missing between the window opening action and window closing action were taken into account. The resulting values were expressed in hours. For the reliable presentation of the classification performance of window states, the evaluation criteria had to take the accuracy of the under-represented labels into account \cite{markovic2017}. In case of the current data set, "open" state is an under-represented class. For that purpose, the relative performance due to imbalanced properties was evaluated using receiver operating characteristic (ROC) and F1 score \cite{fawcett2006}. Resultantly, ROC curves and F1 were included to the metrics proposed by \cite{mahdavi2016}. ROC curves are formed by plotting TP rate over FP rate \cite{he2009}. In addition, the F1 scores are computed as harmonic mean of precision and sensitivity \cite{wolf2017} using Equation \ref{f1}:

\begin{equation} \label{f1}
F1=\dfrac{2TP}{2TP+FP+FN}.
\end{equation}

\section{Experiment setting for additional case-studies}
The performance of the applied modeling approach was evaluated on two additional data sets. The additional evaluation sets consist of offices where at least the variables used for occupant segmentation were available (indoor air temperature, indoor air humidity, indoor $CO_2$ concentration and logged window states). Here, the model trained on original data set, namely E.ON Building, was used as basis. All hyperparameters were fixed after the original hyperparameter search on E.ON Building. Additionally, the activation functions were adapted using the subset of monitoring data from addressed buildings. For this purpose, the additional data set is divided into adaptation set and evaluation set. The size of adaptation set, and the number of iterations were defined based on the changes in the re-learned neuron activations and resulting evaluation performance. The adaptation is conducted in steps of 1000 iterations each, where one iteration is defined as a single forward- and backward pass over a minibatch. The size of the minibatch remained identical to the minibatch size defined by hyperparameter search (4096 data points per batch). 

\subsection{Case-study 1: Model evaluation on an U.S. office building}
With the aim to examine the models' generalization capabilities to different occupants' cultures, building physics and weather conditions, the developed model was evaluated using the data collected on an office building located in Philadelphia, Pennsylvania. Here, the used data set was publicly available and it was originally collected to track the human building interaction \cite{langevin2015}. Similarly to E.ON building, there were operable windows and mechanical ventilation available. The corresponding weather data were downloaded from the NOAA web page \cite{noaa2012}. Since the building’s monitoring data set was logged per 15 minutes, the data points were linearly interpolated to 10 minutes, in order to obtain the same frequency that was used for input features. In total, 21 out of 25 features were available, and around 64k data points were used for evaluation. 

\subsection{Case-study 2: Model evaluation on a sparse data set}

In the scope of this case study, the model's performance was evaluated using a data set, where 8 out of 25 input features were not available. The data were collected over two years of monitoring of a commercial building in Frankfurt, Germany. The monitoring study was conducted by KIT University. For the further details on the monitored building, the reader is referred to \cite{kleber2006} and \cite{schakib2015}. \\
Data was collected by monitoring single- or two-person offices over two years. Logging frequency was ten minutes. Due to the data requirements of the developed model, only the data collected on offices where $CO_{2}$ concentration was measured, were used. As a result, 210k data points from two monitored offices were used for evaluation purposes.
The missing features were approximated linearly using the simplifications presented in Table \ref{tab:kit_missing}. 
\subsection{Case study 3: Application in thermal building simulation}

The developed window states model was incorporated in a calibrated thermal building simulation. For that purpose, a single-zone office model was built using Modelica Dymola and exported as a functional mock up unit (FMU). Eventually, a co-simulation using PyFMI library \cite{andersen2016} was performed. Due to the low availability of building physics data, the simulation was conducted only for the building presented in case study 2.

\FloatBarrier
\begin{table}[ht!]
\centering
   \caption{Applied assumptions used to complement the missing features from the "Frankfurt" data set.  }
     \begin{tabular}{lll}
     \toprule
   Feature & {Unit} & {Complement}  \\
\toprule
     Rain droplets count &  &0.5 \\
    Temperature 100cm  &$^{\circ}$C   &0.8*$T_{outdoor}$ \\
 below ground level &   & \\

  Indoor humidity &  \%  & 30\\
  Set temperatures   & $^{\circ}$C  &23  \\
$T_{1}$ and $T_{2}$   &   &  \\
    Maximal wind speed &  m/s  &1.6*wind speed \\
     Avg. atmospheric pressure &    & 1000\\
     Diffuse radiation &   & 0.3*$I_{global}$\\
     \bottomrule
     \end{tabular}
   \label{tab:kit_missing}
\end{table}  
\FloatBarrier

\begin{figure}[th!]
\centering
 \includegraphics[trim=2cm  0cm 1cm 0cm, width=0.5\textwidth]{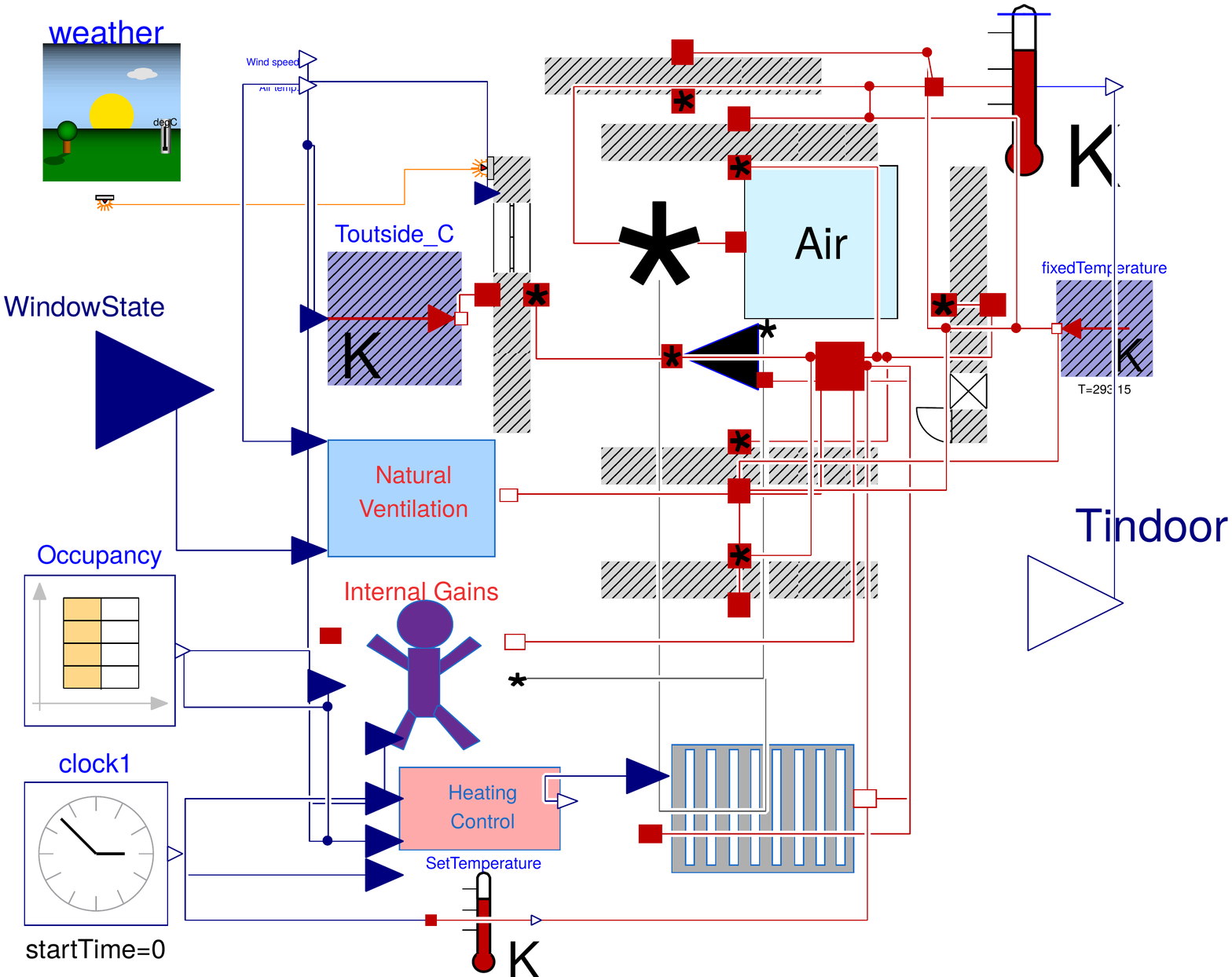}
\caption{Schema of the office's simulation model build in Modelica Dymola.}
\label{fig:model}
\end{figure}

A single zone office (Figure \ref{fig:model}) was modeled using the components from the Aixlib Modelica library \cite{muller2016}. The internal heat gains were defined as occupants and PCs. Heating system was defined as a single radiator, and there was no mechanical ventilation available. Additionally, the temperature of all neighboring rooms was set constant to 20 $^{\circ}$C, and  the ventilation rate was set to \textit{n}=3 [1/h].
\\
Firstly, the simulation model was populated with the measured window states and occupancy profiles in order to quantify the discrepancy between the measured and simulated indoor air temperature. The simulation was conducted for a one-year period in 10 minutes' intervals. The resulting mean squared error was 2.55, while the mean average error was 1.18. 
\FloatBarrier
\FloatBarrier

The simulation input consisted of monitored weather data, monitored occupancy profiles and computed window states using the model that was developed in the scope of this study. The output consists of simulated indoor air temperature, which is used for the identification of the window states on the following time step. The simulations were run using Dymola 2018 in conjunction with Tensorflow 1.8 under Ubuntu operating system.

\section{Results}
\begin{figure}[th!]
\centering
 \includegraphics[trim=1cm  0cm 0cm 2cm, width=0.5\textwidth]{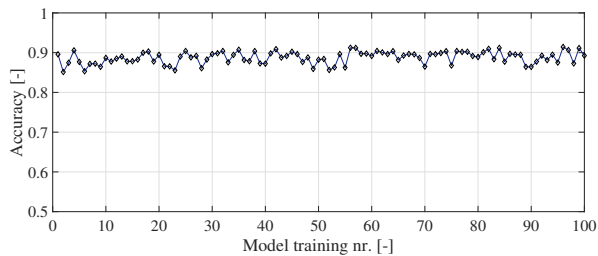}
\caption{Accuracy on the evaluation set (39 offices) for the 100 training procedures with identical network configuration and randomly chosen initial weights.}
\label{fig:acc}
\end{figure}

An optimal performance was achieved for the multi-layer neural network with five hidden layers and the following number of neurons per hidden layer: [64, 94, 81, 10, 25]. The regularization rate was 0.0001 using the L1 norm. It was opted for the adaptive learning rate using a proximal adaptive gradient optimizer and a learning rate equal to 0.1. The chosen activation function was a ReLU. An optimal trade-off between the prediction accuracy and training complexity was scored where the minibatches were implemented with 4096 data points per batch. The model is trained using 10k iterations.
\\
The training and evaluation procedure resulted in approximately 3100 core hours of computation using RWTH compute cluster, with a maximal cache memory usage of 40 GB. Hyperparameter fine-tuning was conducted using a single PC, where the training was conducted using GPU-based computations. The GPU work memory usage was around 15 GB, which corresponded to around 40\% of the working memory available of the used graphics processor, while the clock time needed for 1000 iterations during training procedure corresponds to approximately 0.8 seconds (clock time). 
\\
Since the initial weights were randomly chosen in order to avoid a model bias, it was opted for sequential training of 100 separate models using the identical configuration. Here, the final model accuracy was evaluated as the mean value of the results scored during each model training. The single accuracy scores for each training procedure are presented in Figure \ref{fig:acc}.\\

\FloatBarrier
\FloatBarrier

\begin{figure}[th!]
\centering
 \includegraphics[trim=0cm  2cm 0cm 0cm, width=0.45\textwidth]{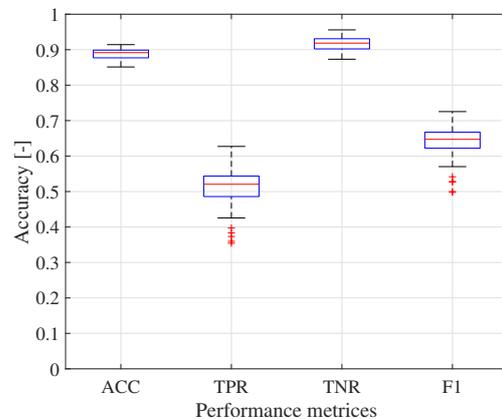}
\caption{Accuracy, true positive rate and true negative rate on the evaluation test for the sequentially separated training of 100 models using identical neural network configuration and randomly generated initial weights.}
\label{fig:acc_tp_tn}
\end{figure}

\begin{figure}[h!]
\centering
 \includegraphics[trim=0cm 0cm 0cm 0cm, width=0.45\textwidth]{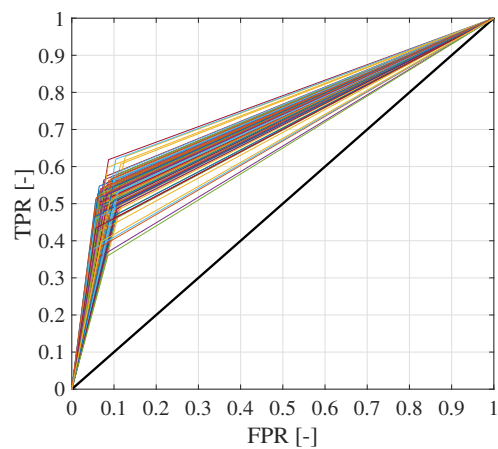}
\caption{ROC diagrams for the 100 models with random initial weights.}
\label{fig:scatter_tp_tn}
\end{figure}

The statistical analysis of the absolute evaluation results for the 100 repeated training procedures in terms of TPR, TNR, accuracy and F1 are presented in Figure \ref{fig:acc_tp_tn} and Table \ref{tab:acc_tp_tn}. Since the quantiles, mean and median values of the performance metrics showed low variance (between 0.02 and 0.05 for each metric), the model was not over fitted to the training set, and the network is stabilized with the respect to the random initial weights. The resulting mean accuracy for the training of 100 models was 89\%. The TP and TN rates are 52\% and 92\% respectively. The relative performance of the proposed method is presented graphically using an ROC diagram (Figure \ref{fig:scatter_tp_tn}). As presented in Figure \ref{fig:scatter_tp_tn}, the trained model in case of all 100 random initial weight guesses performed better than a random guess (diagonal line). Also, it may be interpreted that the trade-off between the improvement in the TPR and FPR remained rather low. Additionally, the deviation in the TPR score of each individual classifier was not caused by the variance of FPR. Rather, it may be interpreted as the cause of the "goodness" of the initial weights in a few out of 100 repeated model trainings, while the FPR remains high due to data imbalance.

\FloatBarrier
\begin{table}[ht!]
\centering
   \caption{Prediction performance of the investigated models for window opening. }
     \begin{tabular}{lllll}
\toprule
&ACC &TPR  & TNR & F1\\
&[-] &[-] & [-] & [-]\\
\toprule
min	&	0.85	&	0.35	&	0.87	&	0.50	\\
25 \% quantile	&	0.88	&	0.49	&	0.90	&	0.62	\\
mean	&	0.89	&	0.51	&	0.92	&	0.64	\\
median	&	0.89	&	0.52	&	0.92	&	0.65	\\
75 \% quantile	&	0.90	&	0.54	&	0.93	&	0.67	\\
max	&	0.91	&	0.63	&	0.96	&	0.73	\\

 \bottomrule
     \end{tabular}
   \label{tab:acc_tp_tn}
\end{table}  
\FloatBarrier

\FloatBarrier
\FloatBarrier

\FloatBarrier
\FloatBarrier

\subsection{Case study results}

\begin{figure}[th!]
\centering
 \includegraphics[trim=1cm  0cm 0cm 0cm, width=0.5\textwidth]{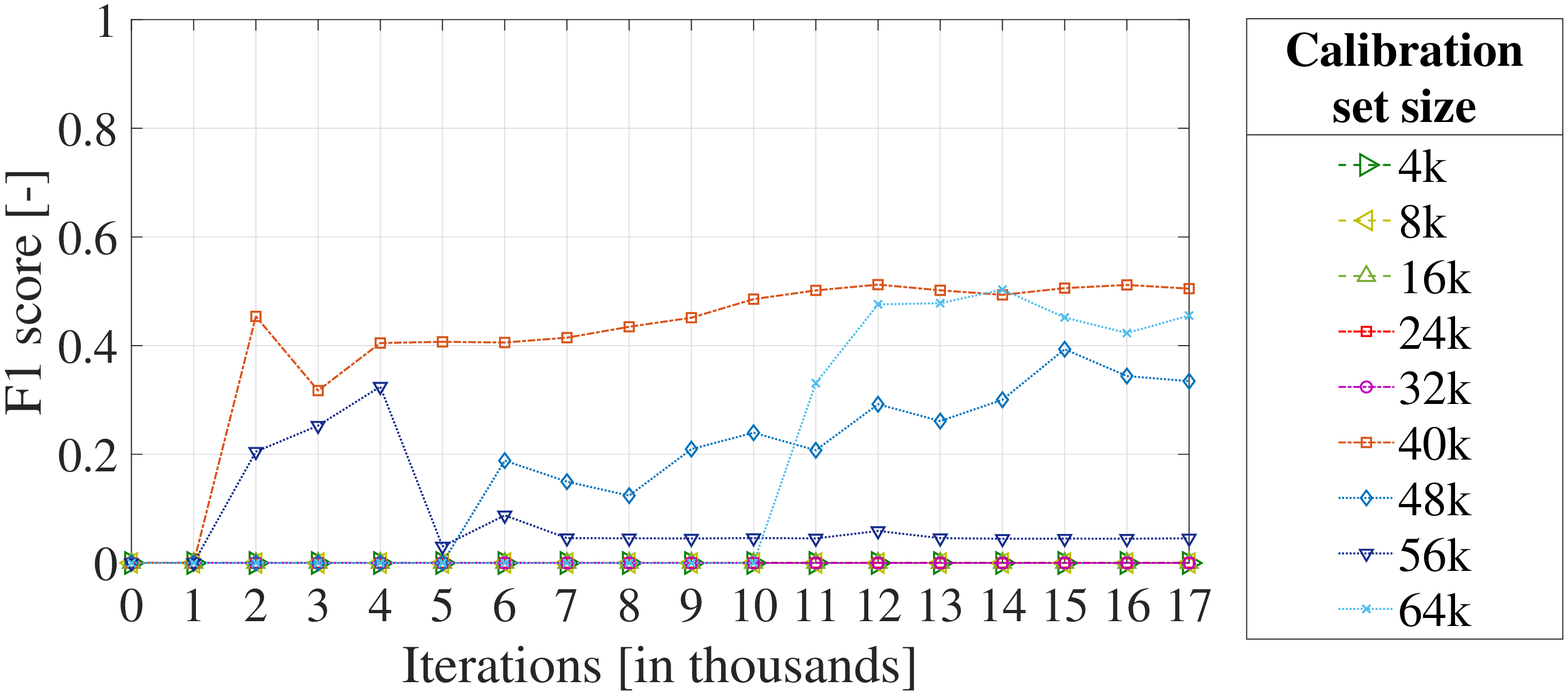}
\caption{Evaluation performance in terms of F1 score as the function of learning iterations over varied adaptation set size for the building in Philadelphia (case study 1).}
\label{fig:adapt_kit}
\end{figure}

\begin{figure}[h!]
\centering
 \includegraphics[trim=1cm  0cm 0cm 0cm, width=0.5\textwidth]{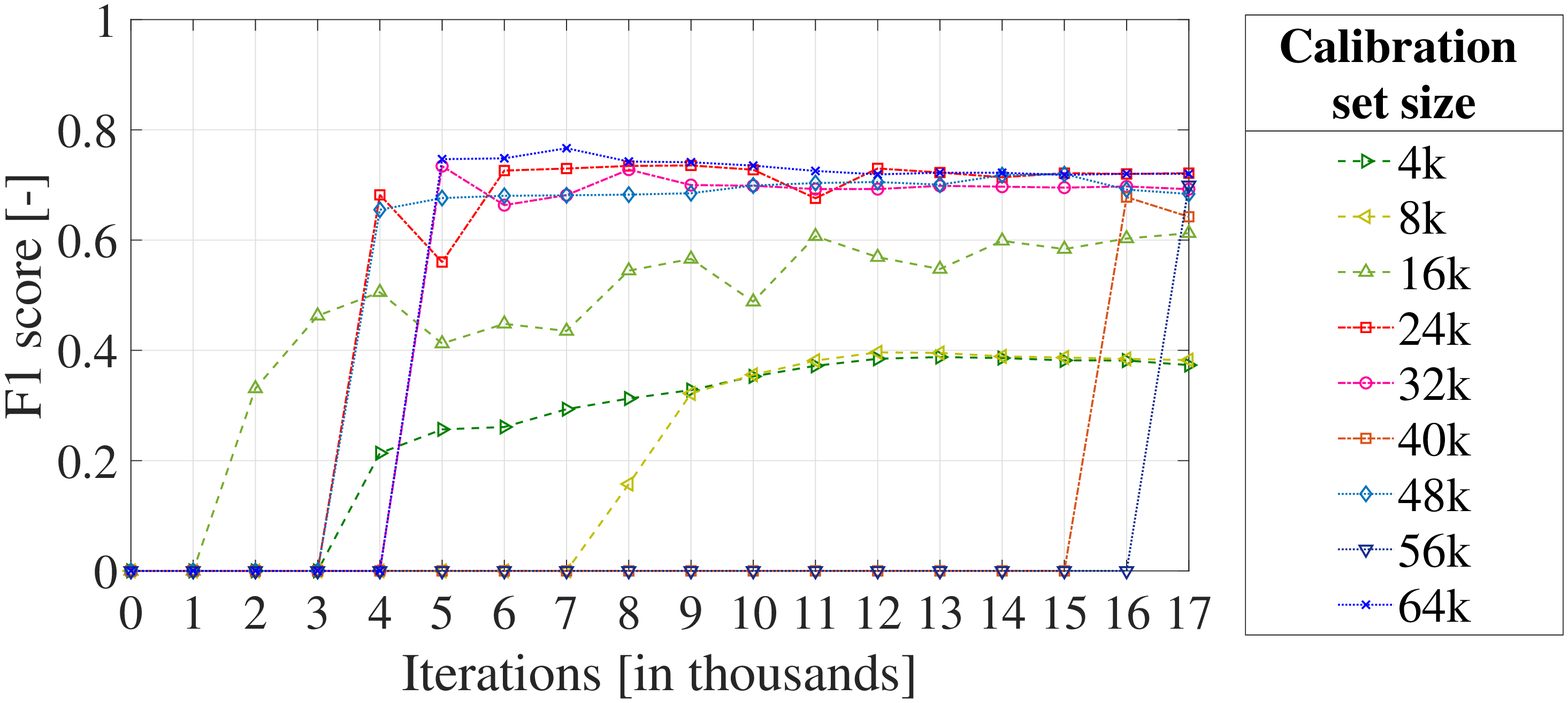}
\caption{Evaluation performance in terms of F1 score as the function of learning iterations over varied adaptation set size for the building in Frankfurt (case study 2).}
\label{fig:adapt_us}
\end{figure}

The weights were adapted by executing additional iterations over the "adaptation subset" of the subset, while all parameters remained identical. The resulting performance is presented in Table 4 and Figure 11. Additionally, the performance was presented in Figures \ref{fig:adapt_us} and \ref{fig:adapt_kit} as the function of learning iterations in batch mode. In case of the office building from the case study 1, activations were adapted by running 12k iterations over 40k sequential data points from the addressed data set. Using the batch size defined in previous sections, this resulted in 1200 adaptation epochs. In case of the Frankfurt data set, adaptation is conducted during 8k iterations over 24k data points, which resulted in 1670 training epochs. Resultantly, the evaluation performance is evaluated using the rest of the data: approximately 24k data points from case study 1 and 185k data points from case study 2.

\FloatBarrier
\FloatBarrier

\FloatBarrier
\FloatBarrier

The absolute performance results are summarized in Table \ref{tab:actions_duration}. In case of the E.ON ERC building, the fraction of time where windows were opened was slightly overestimated (4 percent points). The median duration of sequences where status was "open" was underestimated for 0.31 h, while the predicted number of actions per day were larger, when compared to the monitoring data. Resultantly, the predicted sequences where the windows were closed are shorter, compared to the monitoring data. Evaluation results on the case studies 1\footnote{The data set from the "Case study 1" contained an interval, where windows remained opened between September 30th, 2012 and October 24th, 2012. As a consequence of this long interval where the windows remained opened, the results of the interquartiles and median values were screwed towards a longer duration of sequences with opened windows. Due to that, the descriptive statistics of opening durations was additionally presented for the case, where the data regarding this specific sequence were excluded.} and 2 pointed out similar model's tendency- the sequences where the window status was unchanged were shorter, compared to the monitoring data, and the actions per day were overestimated for the Frankfurt data set (case study 2).

\FloatBarrier
\begin{table}[ht!]
\centering
   \caption{Prediction performance of the investigated models for window opening. }
     \begin{tabular}{lllll}
\toprule
&ACC &TPR  & TNR & F1\\
&[-] &[-] & [-] & [-]\\
\toprule
Case study 1 &	0.86&0.37   &0.96	&0.53	\\
Case study 2 &0.73	&0.76&	0.71&	0.74	\\
Case study 3 &	0.63 &0.85 & 0.55  &		0.74\\

 \bottomrule
     \end{tabular}
   \label{tab:kit_acc_tp_tn}
\end{table}  
\FloatBarrier
\FloatBarrier
\begin{figure}[ht!]
\centering
 \includegraphics[trim=0cm  0cm 0cm 0cm, width=0.50\textwidth]{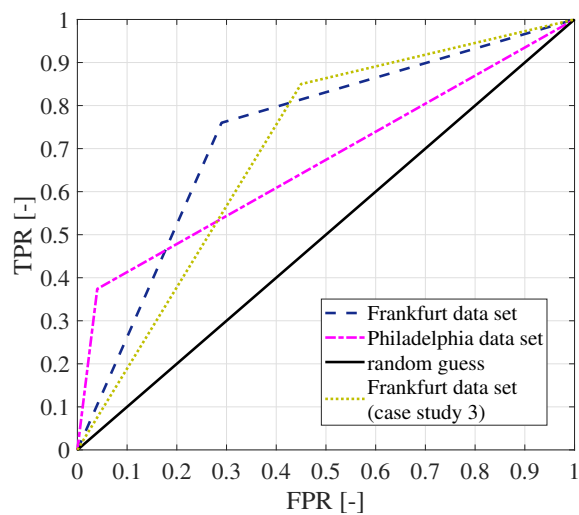}
\caption{ROC diagrams for the identified window states in case of buildings from the case studies 1 and 2.}
\label{fig:roc_kit}
\end{figure}

\begin{table*}[th!]

\centering
   \caption{ Absolute results obtained by the trained model for the E.ON ERC building, case study 1, case study 2 and in case of incorporation of the co-simulation (case study 3).}

     \begin{tabular}{lllllllllll}
     \toprule
     \multicolumn{2}{l}{} &  & \multicolumn{4}{l} {Opening duration} & \multicolumn{4}{l} { Closing duration}    \\
          \bottomrule
	&open state 	&actions 	&	25 \%  	&	median	&	75 \%  & IQR 	&	25 \% 	&	median	&	75 \%  & IQR  	\\
	&		&	per day	&	quant. 		&		& quant.  &		&		quant.	&		&	quant.	&	\\
&	[-]  &	[1/d]	&	[hrs]	&	[hrs]	&	[hrs]	&[hrs]	&	[hrs]	&	[hrs]	&	[hrs]	&	[hrs]	\\
     \bottomrule
    \multicolumn{5}{l}{E.ON ERC Data Set	}  				&	&  &	&		&		&		\\
         \bottomrule
observed &	0.07&	1.02		&	0.12	&	0.47	&	1.48	&1.36		&	1.57	&	7.7	&	21.58& 20.01	\\
predicted &	0.11&	2.53	&	0.06	&	0.18	&	0.75	&0.69	&	0.08	&	0.352	&	4.54& 4.46	\\
     \bottomrule
    \multicolumn{5}{l}{Case study1: Philadelphia data set 	}  		&&		&		&		&		&		\\
         \bottomrule
observed&0.15	&	0.02		&	2.00	&	27.67	&	309.42	&307.42	&	23.08	&	70.08	&	155.17 & 132.09		\\
observed$^2$& &		&	1.83	&	2.33	&	40.33	& 38.70 &		&		&	 & 		\\

predicted&0.10	&	0.60		&	0.33	&	0.83	&	3.33	&3.00	&	0.33	&	1.17	&	4.00 & 3.67		\\
     \bottomrule
    \multicolumn{5}{l}{Case study 2: Frankfurt data set 	}  		&&			&		&		&		\\

         \bottomrule
observed&	0.35&	1.37		&	0.17	&	0.67	&	4.17	&4.00	&	1.17	&	2.83	&	5.67	& 4.50	\\
predicted&0.48	&	2.04		&	0.33	&	0.67	&	2.63&2.30	&	0.33	&	0.50	&	1.83	& 1.50	\\
     \bottomrule
    \multicolumn{5}{l}{Case study 3: One-year simulation data}  		&&		&		&		&		&		\\
         \bottomrule
observed	&	0.26	&	2.86	&	1.17	&			2.83	&	7.00	&	5.83&	0.17	&			0.33	&	1.50	& 1.33	\\
predicted	&	0.56	&	2.55	&	0.33	&			0.67	&	2.00	&1.67	&	0.33	&			1.00	&	3.50	& 3.17\\

     \bottomrule
     \end{tabular}
   \label{tab:actions_duration}
\end{table*}  
\FloatBarrier

\section{Discussion}


The occupant segmentation was performed with the aim to identify the subset of data for an efficient algorithmic implementation of the identification the window states. However, the segmentation and clustering results do not refer to the actual physiological features of the investigated users, but to the measured data used in the further modeling. Similarly, to the results presented by \cite{obrien2017}, the available sample size in this study (52 offices) is not sufficient to raise conclusions about the large scale occupant segmentation. For instance, large scale would be considered data collected in climate zones that were not addressed in the scope of this study or occupants with different cultural background. 

The performance results were higher in case of multiple hidden layers, when compared to cases where a single hidden layer was implemented. This is a result of a learned feature mapping \cite{lecun2015} over multiple hidden layers. Hence, the narrow architecture of the optimal network structure (5 hidden layers) indicates that a good performance may be achieved with a low model complexity, compared to the modeling needed for most of the vision- and speech- recognition tasks. The model performance was improved where both data from the current and the previous time steps were used as inputs. The need for the information from the previous time steps is caused by the sequential nature of the OB actions, which shows the need for the inclusion of the time series modeling of OB actions. 


In contrast to the related studies, the model is evaluated with the unseen occupants from the data set in question, instead of the unseen data collected on the same occupants as the training set. Evaluation results for the case of E.ON ERC building pointed out that the model could identify the window states with 89 \% accuracy, 52 \% TPR and resulting median F1 score of 0.65. The absolute results showed similar performance in terms of overall accuracy in the case of a mechanically air-conditioned office building in Philadelphia, while the TPR dropped to 37 \%. In the scope of second case study, the model was evaluated on an office building in Frankfurt, where approximately 30 \% of the input features were missing. In this case, the model performed less accurate in terms of overall accuracy and TNR, although the accuracy of correctly identified window states was higher than in the case of the original building. 

\begin{figure}[th!]
\centering
 \includegraphics[trim=2cm  1cm 2cm 1cm, width=0.4\textwidth]{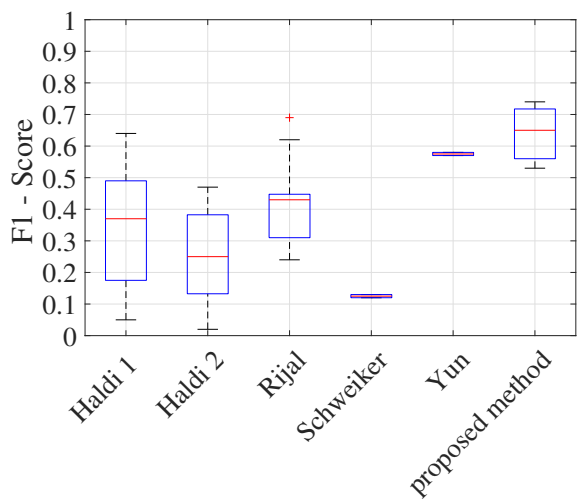}
\caption{Comparison of F1 score for the building-level calibrated window opening models that were evaluated using data from the buildings different from the original building. Compared models refer to the following methods from the related literature: Haldi1- logistic regression \cite{haldi2009}, Haldi2- Markov model \cite{haldi2009}, Rijal- \cite{rijal2007}, Schweiker-\cite{schweiker2009}, Yun- \cite{yun2008}. }
\label{fig:comparison_models}
\end{figure}

The proposed model's performance in terms of handling the data imbalance in combination with overall accuracy was analyzed using the F1 score. The F1 score of the proposed model for the three evaluation sets were compared to the performance reported in the number of related studies (\cite{haldi2009}, \cite{haldi2013}, \cite{langevin2015}, \cite{laurent2017}, \cite{mahdavi2016}, \cite{rijal2007}, \cite{schweiker2009}, \cite{wolf2017}) that were calibrated on the building level (Figure \ref{fig:comparison_models}). The considered related studies were publications where the original models (x-axis) were evaluated in original form, evaluated in calibrated form, or modified and calibrated by a data set different from the original set. The results pointed out, that the proposed method showed higher F1 score, when compared to alternative modeling approaches. However, the results of the proposed method were obtained from the internal study in the scope of same publication. In order to raise the conclusions regarding the model's applicability and generalization capabilities, additional evaluation in terms of round robin studies or double blind studies are required. Additionally, the variance of results of each related study may be carefully interpreted. For instance, the variance for models proposed by Schweiker \cite{schweiker2009} and Yun \cite{yun2008} is significantly lower, when compared to alternative models' variance. Hence, the lower variance may be the result of the smaller number of studies where the original model was evaluated. Resultantly, no conclusion regarding the variance in performance of the proposed method and its' comparison to the alternative modeling approaches can be drawn. 
\FloatBarrier

\FloatBarrier

The relative results pointed out that the proportion of time where windows were opened and the number of actions per day were slightly overestimated for all three data sets, while the durations of the sequences where windows were opened were predicted to be shorter, compared to the monitoring data. Resultantly, the duration of sequences with closed windows were predicted to be shorter than logged in the monitoring data.  

Currently, the model is evaluated using data from the offices in Aachen, Germany, data set collected in Philadelphia, USA and offices in Frankfurt, Germany. Additional differences between the data sets in question come from the buildings' architecture, sensor equipment, sensor positions inside the offices and the availability of mechanical air conditioning. Nonetheless, additional experiments and evaluation using the data from different types of buildings and different climate zones would be beneficial to quantify the performance for further building types and climates.


The model did not require any parameter search or calibration, when applied on the additional buildings and the only required step was weight adaptation. This is conducted by running several epochs of training using a subset of “adaptation data set” from an additional building. Resultantly, the model incorporated the specific features of the building in question to the already learnt knowledge from a significantly larger training set. Additionally, it "forgot" some characteristics of the original building. Since the window opening model is rather low-dimensional, a model could possibly be further developed where no adaptation is required. For this purpose, the data from additional sensor sources should be incorporated as model input (for instance type of the building, physiological- and comfort data). Another crucial point for the transfer of the model to different buildings and occupants is the scaling of the features, namely scale adaptation. Besides the limits of the scaling, it is necessary to incorporate the distribution and frequencies of the measured values in the scale metrics, in order to reflect the actual drivers of the window openings reliably. There is little research done on the scale adaptation of the OB data in buildings, and it is still an open question. 
\\
The model is incorporated in a Dymola-based thermal simulation with feedback using PyFMI. The absolute and relative performance results showed that the model tends to overestimate the proportion of time where windows were opened. There are several open research questions that need to be addressed in order to make the full use of this modeling approach. Firstly, one of the model's inputs is the indoor $CO_{2}$ concentration. Hence, the simulation of the indoor $CO_{2}$ concentration requires a fluid dynamics approach, which is not available in the conventional building simulations tools. Secondly, the window states were evaluated in 10 minutes frequency, which resulted in high communication costs between FMU and developed model. Here the communication cost may be reduced by finding the optimal frequency at which the window state can be evaluated while the accuracy would remain in similar range. 
\\
The model used 22 variables from the current timestep and 3 variables from the previous timestep as input features. Alternatively, the same model input could be addressed using a recurrent neural network (RNN) to learn the temporal dependencies. In that case, the weather data could be addressed using a feed-forward neural network, while the indoor climate would be addressed using an RNN, where the back-propagation through time (BPTT) would be added on the input layer of the indoor climate data. In this particular case, where only the indoor climate data from one previous time step was used as input feature, the use of RNN would lead to higher model complexity and no significant accuracy improvement, since the temporal dependencies were already addressed by unrolling the sequence into a single input layer of a feed-forward neural network. Nonetheless, RNNs, including gated structures such as long-short-term memory (LSTM) and leaky units, are a very promising approach for modeling longer time-series of OB in buildings. In particular, future research should address the optimal network architecture, duration of the input time-series and the temporal discretization for the RNNs.

\section{Conclusion}
This work presented a window opening model based on deep learning methods. The developed model led to a significant improvement regarding modeling imbalanced properties of window states, when compared to alternative modeling approaches. Additional advantage of the proposed deep learning-driven OB modeling approach were satisfying generalization capabilities and robustness towards individual occupants. Application of multiple hidden layers in neural network was beneficial for improving the accuracy, while the overall model complexity remained low. The practical applicability of the proposed method is evaluated in the scope of three case studies. Their results pointed out competitive results, when compared to alternative building-wise calibrated window opening models.  
\\
In order to make the developed method accessible, the trained model and the Python-based scripts for model evaluation and weight adaptation will be published as open source repository. Resultantly, the model may be used by the engineers and designers as standalone, or incorporated in thermal building simulation.

\section{Acknowledgements}
The authors appreciate the financial support of this work by the German Federal Ministry of Economics and Energy (BMWi) as per resolution of the German Parliament under the funding code 03ET1289D. We thank Mark Wesseling and Davide Cali from EBC Institute, E.ON ERC at RWTH Aachen for providing the monitoring data, as well as to the Physical Geography and Climatology Group for providing weather data. Also, we thank Marcel Schweiker and Andreas Wagner of Karlsruhe Institute of Technology, Germany, for sharing the data set "KfW Ostarkade", which was used for the model validation.

\clearpage
\begin{Large}
Supplementary Material 
\end{Large}

\section*{Overview}
This supplementary material contains additional information to the work entitled “Window Opening Model using Deep Learning Methods”. It contains the following sections:
\begin{itemize}
\item {Data set – additional information on the performed data preprocessing and the used data set}
\item {Occupant segmentation – generated visualization maps of the occupant segmentation results}
\item {Implementation details – further details on model implementation, that would be particularly beneficial for model reproduction}
\end{itemize}

\appendix

\section{Data set}
\subsection{E.ON ERC Data set}
Due to data loss, not every office provided monthly monitoring data (Figure \ref{fig:ebc_data}(a)). In addition, data points where window position information was missing were excluded from the further evaluation.  As a result, there were approximately 19 mio. data points used for further modeling and evaluation. An overview of the amount of data points for each monitored office is shown in Figure \ref{fig:ebc_data}(b).
\\
Weather data were measured at a weather station on the university campus, which is located 1.34 km air distance from the monitored building. The weather data acquisition is conducted by the Physical Geography and Climatology Group at RWTH Aachen University. For more information on the data collection procedure, the reader is referred to \cite{schneider2014} and \cite{schneider2015}.
\\
The minute-wise logged building's monitoring data were joined with corresponding weather data logged in 10 minutes' frequency, so that each data point from the weather data set was joined with the previous five or the following five minutes of the indoor climate data.
\FloatBarrier
\begin{figure}[h!]
\centering
 \includegraphics[trim=2cm  0cm 12cm 0 cm, width=0.5\textwidth]{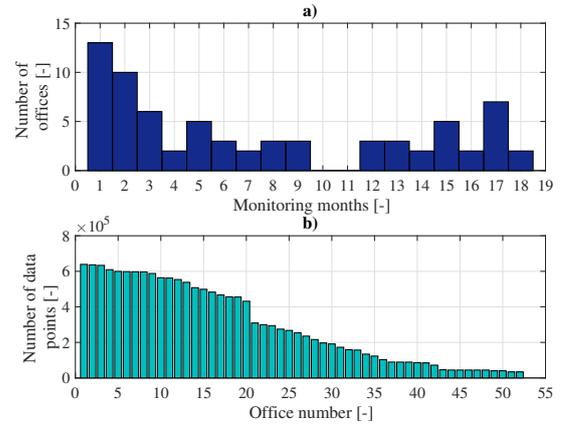}
\caption{Histogram representing the number of monitoring months available (x-axis) for each office (y-axis) (a); Amount of data points representing all necessary features for each used office (b). }
\label{fig:ebc_data}
\end{figure}
\FloatBarrier

\textcolor{black}{There were two operable windows in each monitored office. The window states were logged as a binary variable based on the state of both windows. Here, the window state was defined as “opened” in case at least one of the two windows was opened. Additionally, it was not distinguished between different positions of opened windows (for example tilted/completely opened). 
\\
Although there could be a certain proportion of occupants that changed their workplace during the monitoring study (mostly due to work contract expiration or a new employment), this effect was not considered in the scope of this study due to the lack of ground truth data. Consequently, it was not distinguished between the actions performed by different occupants on the same windows.}
\\

Feature scaling was performed in order to avoid numerical problems during training and evaluation procedures. Since the model evaluation was performed using an independent data set whose limits were not known prior to the training procedure, a unique scale needed to be adopted. The limits were set based on the empirical measurements and on the basis of the theoretical values for climate data (Table \ref{tab:data_ebc}). The binary variables used include occupancy and window position, where "1" indicates occupant's presence and opened window, respectively. The temporal data include Unix timestamps (between 2002 and 2022), hour of the day (between 0 and 23) and day of the week (between o and 6, where 0 corresponds to Monday).

The indoor set point temperature [$^{\circ}$C] indicates the manually adjusted air temperature for air conditioning, which was scaled in the range of the comfort zone between 18 and 26$^{\circ}$C. The indoor air temperature [$^{\circ}$C] was scaled within the same range as the outdoor temperature, in order to cover the potential cases where the window remained opened over a longer period of time. The indoor relative humidity [\%] was scaled in range from 0 to 100 \%. The indoor $CO_{2}$ [ppm] concentration was assumed to be between 0 and 2500 ppm for single and double occupied offices, based on measured values.  \\
The outdoor temperatures [$^{\circ}$C] were scaled for the plausible range for the continental climate in Germany. These include the outdoor air temperature from the nearby weather station, ground temperature in order to depict the climate from the previous days and the temperature measured on the outside facade of each office. The outdoor humidity is in the range from 0 to 100\%. The number of rain droplets and volume are scaled based on the measured values from the weather station on RWTH Campus. 
\\The global radiation is scaled between 0 and 1365 $\dfrac{W}{m^2}$. The latter describes the amount of solar radiation received at the top of the atmosphere on a normal plane at the mean Earth-Sun distance \cite{nasa2017}. The diffuse radiation is scaled between 0 and 800 $\dfrac{W}{m^2}$. The average pressure [mbar] for Aachen is scaled between 900 and 1100, based on measured values. 
The wind direction was scaled between 0$^{\circ}$ and 360$^{\circ}$, where 0$^{\circ}$ indicates south wind. The wind speed and maximal wind speed were scaled between 0 and 28.61 m/s, where the upper bound indicates wind class "10" \cite{weather2013}.
\subsection{Data set from the case study 1}
Here, the used data set was publicly available and it was originally collected for tracking the human building interaction \cite{langevin2015}. The used data were collected on two occupants where the measured values of $CO_{2}$ concentration and the window states were available. This include to data point range between approximately 245200-274000 and 700800-728700 from the original data set. Additionally, the data point that contained “NaN” values were excluded from the evaluation. In total, there were around 40k interpolated data points used for weight adaptation (26k data points from the original data set) and 24k interpolated data points used for model evaluation (16k data points from the original data set).
\newpage
\section{Occupant segmentation}

\FloatBarrier
\begin{figure}[h!]
\centering
 \includegraphics[trim=0cm  4cm 0cm 0 cm, width=0.5\textwidth]{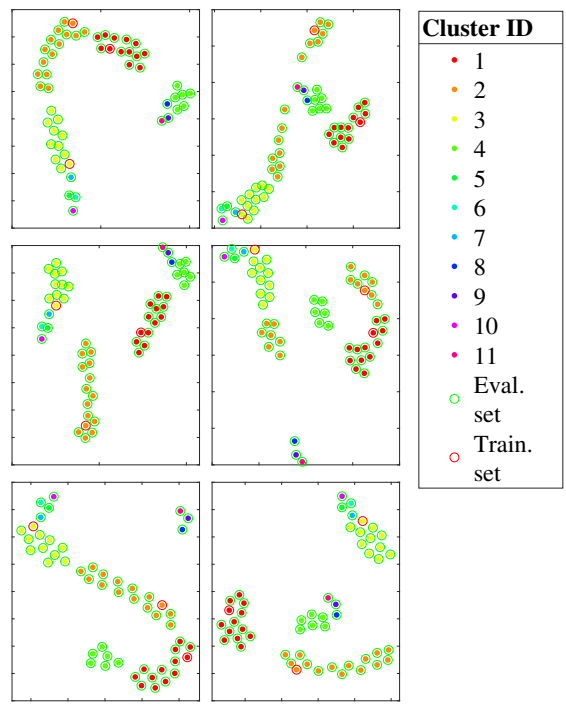}
\caption{Repeatedly generated t-SNE visualization maps of the occupant segmentation results using same parameters. The perplexity had a constant value of "4” for the generation of all plots. }
\label{fig:ebc_data}
\end{figure}
\FloatBarrier

\section{Implementation details}
\subsection{Batch and minibatch choice}
\begin{figure}[ht!]
\centering
 \includegraphics[trim=0cm  0cm 0cm 0cm, width=0.45\textwidth]{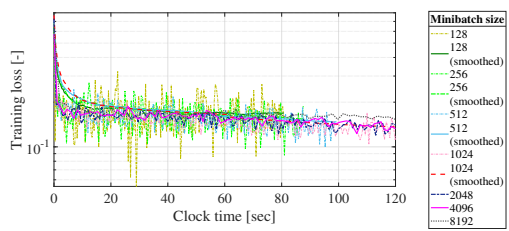}
\caption{Relationship between the model training progress and the required clock time.}
\label{fig:minibatch}
\end{figure}

The computation of deterministic gradient descent is computationally expensive \cite{goodfellow2017} and not feasible in case of a complex data set with a large amount of training data samples, which was the case for modeling the window states using the given training set. Hence, the chosen training method was the minibatch stochastic method. 
\\
 As a part of the minibatch method, a small subset of the training data was used to train the model. Consequently, the final model was the result of the average of all the sub models that were developed using the minibatches of the training data. The following criteria were taken into account when choosing the size of the minibatches: effective use of the multicore processor used for the computations and sensitivity of the applied algorithm with respect to the sampling error. The range of the tested minibatch sizes was made based on the following two findings of the minibatch training presented by \cite{goodfellow2017} - firstly, the applied multicore architectures are underutilized by extremely small batches. Secondly, in this study, the updates were computed using the first order gradient, which is robust \textcolor{black}{to sampling error} compared to the second order methods, for instance Hessian. As a result, the examined minibatch size remained relatively low, ranging between 128 and 9192 data points per batch.
\\
\textcolor{black}{The minibatch size was analyzed in order to find an optimal combination of following criteria: efficient use of local multicore processors used for training, reducing the clock time required for model training and minimizing the cross-entropy training loss. For that purpose, a set of experiments was conducted in order to qualitatively evaluate the impact of the minibatch size on the neural network training progress and the results are presented in Figure \ref{fig:minibatch}. Here, a neural network was tuned where the hyperparameters were in the range of tuned values, with 4 hidden layers \`a 50 neurons, learning rate 0.1 and regularization coefficient of 0.1. The minibatch size was varied between $2^{7}$ and $2^{13}$ data points and the resulting training loss values are presented in figure \ref{fig:minibatch}. Here, it was opted for power of 2, in order to achieve better runtime on distributed architectures.
\\
Defining an efficient training procedure is commonly multi objective optimization, since the optimal cores usage do not necessarily result in minimal clock time \cite{goodfellow2017}. Resultantly, it was opted for 4096 data points per minibatch, due to the satisfying combination of multicore usage and required clock time. Additionally, the training loss for the chosen minibatch size was lower, when compared to alternative minibatch sizes.
}
\FloatBarrier

\subsection{Neural network architecture design}
The number of hidden layers and the number of neurons for each hidden layer were treated as hyperparameters that were optimized with respect to the prediction accuracy. The number of hidden layers was investigated in range between 1 and 7, while the number of neurons was varied between 10 and 100 per each hidden layer. 
An optimal number of neurons for a narrow network ( $\leq$ 3 hidden layers) was searched using a grid search method, while the number of neurons for deeper architectures was investigated using the approach initially presented by \cite{Bergstra2012}, with 500 random training choices for each combination of number of hidden layers and batch size.
\\
 As a part of hyperparameter search, different number available of neurons per hidden layer were investigated. Additionally, the introduction of regularization penalty, a portion of weights was set to zero, which may result in excluding the impact of all weights from a certain neuron on the prediction. Consequently, the number of neurons was estimated recursively with respect to all other hyperparameters. As a result, the number of neurons did not have an explicit impact on the predictive performance. Rather, the optimal combination of the number of neurons with additional hyperparameters impacted the resulting predictive performance.
 \\
\textcolor{black}{The relationship between the number of hidden layers and the model's predictive performance was analyzed. For that purpose, the optimal hyperparameter combination for each number of hidden layers was identified. The average results over 20-30 repeated training procedures are summarized in Table \ref{tab:acc_layers}. Given a suitable combination of hyperparameters, namely learning rate, regularization, number of training iterations and sufficient number of neurons per hidden layer, the performance was robust towards the varied number of hidden layers in the range between 4 and 7. Based on the predictive performance, it could be observed that the performance on the over-represented class was similar for different number of hidden layers. However, the performance on the under-represented class (open windows) varied with respect to the number of hidden layers. A model with five hidden layers was identified to have an optimal predictive performance\footnote{\textcolor{black}{The performance of the four layer network was only marginally lower than the performance for the case of five hidden layers. However, a neural network with four layers did not result in significantly lower model complexity, since it required higher regularization penalty. In addition, the number of neurons were in the same range as it was in the case of five hidden layers.}}.}

\FloatBarrier
\begin{table}[ht!]
\color{black}
\centering
   \caption{Predictive performance for the models with a varied number of hidden layers. }
     \begin{tabular}{lllll }
\toprule
number of 	&	ACC 	&	TPR	&	TNR	&	F1	\\
hidden layers	&	[-]	&	[-]	&	[-]	&	[-]	\\
\toprule
1	&		0.89&	0.38	&	0.93	&	0.53	\\
2	&	0.89	&	0.45	&	0.92	&	0.59	\\
3	&	0.89	&	0.37	&	0.94	&	0.51	\\
4	&	0.89	&	0.50	&	0.92	&	0.63	\\
5	&	0.89	&	0.51	&	0.92	&	0.64	\\
6	&	0.89	&	0.49	&	0.92	&	0.62	\\
7	&	0.89	&	0.51	&	0.92	&	0.64	\\
 \bottomrule
     \end{tabular}
   \label{tab:acc_layers}
\end{table}  
\FloatBarrier
\subsection{Learning rate and regularization}
In order to avoid overfitting, a regularization parameter is introduced. Regularization is any modification we make to a learning algorithm that is intended to reduce its generalization error but not its training error and it is necessary to choose a form of regularization that is well-suited to the particular task in question \cite{goodfellow2017}. Based on the small size of the trained network in terms of number of features and hidden layers, it is opted for the weight shrinking using "\textit{L1}" regularization. An optimal regularization score was searched in a range between 0.00001 and 0.9. 
\\
In addition, a gradient descent-based learning rate was chosen. It was opted for the proximal adaptive gradient optimizer, due to its adaptive step possibilities (available as a part of the Tensorflow library\cite{abadi2016}). The tuning was conducted for a learning rate in the range between 0.01 and 0.1. The tested activation functions were rectified linear functions and hyperbolic tangents.
\subsection{Further hyperparameters}
The maximal number of iteration was set based on the convergence criteria of the loss function. Eventually, it was opted for 10k iterations during the model training. At that point no significant advances in learning process occurred, which resulted in converged loss function and no significant changes in activations.

\subsection{Computational environment}
The model was developed using the Tensorflow \cite{abadi2016} library for Python 3.6. The hyperparameters were tuned using computational resources from RWTH compute cluster (CPU only) and a personal computer (combined GPU and CPU-based computations). The personal computer was running on Ubuntu 16.4 operating system. The processor used was Intel Core i7-6900K (3.2 GHz), while the used GPU is Nvidia GeForce GTX 1080.

\end{document}